\documentclass[10pt,twocolumn,letterpaper]{article}

\usepackage[final]{cvpr}      

\usepackage{xcolor}
\usepackage{stfloats}
\usepackage[most]{tcolorbox} 
\definecolor{boxgray}{HTML}{F9F9FB}   %
\definecolor{mblue}{HTML}{1A4C9C} 
\definecolor{titleblue}{HTML}{2E5FA3} %
\definecolor{frameblue}{HTML}{C6D6F2} %

\tcbset{
    colback=boxgray,
    colframe=frameblue,
    coltitle=titleblue,
    boxrule=0.5pt,
    arc=1mm,
    left=2mm,
    right=2mm,
    top=1mm,
    bottom=1mm,
    fonttitle=\bfseries\small,
    width=\linewidth
}
\usepackage[normalem]{ulem} 
\newcommand{\best}[1]{\textbf{\textcolor{red}{#1}}}
\newcommand{\secondbest}[1]{\textcolor{blue}{\uline{#1}}}
\usepackage{multirow}
\definecolor{cvprblue}{rgb}{0.21,0.49,0.74}
\usepackage[pagebackref,breaklinks,colorlinks,allcolors=cvprblue]{hyperref}

\def\paperID{*****} 
\def\confName{CVPR}
\def\confYear{2026}

\title{Guiding Perception-Reasoning Closer to Human in Blind Image Quality Assessment}

\author{
    Yuan Li \quad Yahan Yu \quad Youyuan Lin \quad Yong-Hao Yang \quad Chenhui Chu \quad Shin'ya Nishida \\
    Kyoto University \\
    {\tt\small li.yuan.67n@st.kyoto-u.ac.jp}
}

\begin{document}
\twocolumn[{%
\renewcommand\twocolumn[1][]{#1}%
\maketitle
\begin{center}
    \centering
    \vspace{-2.3em}
    \includegraphics[width=\linewidth]{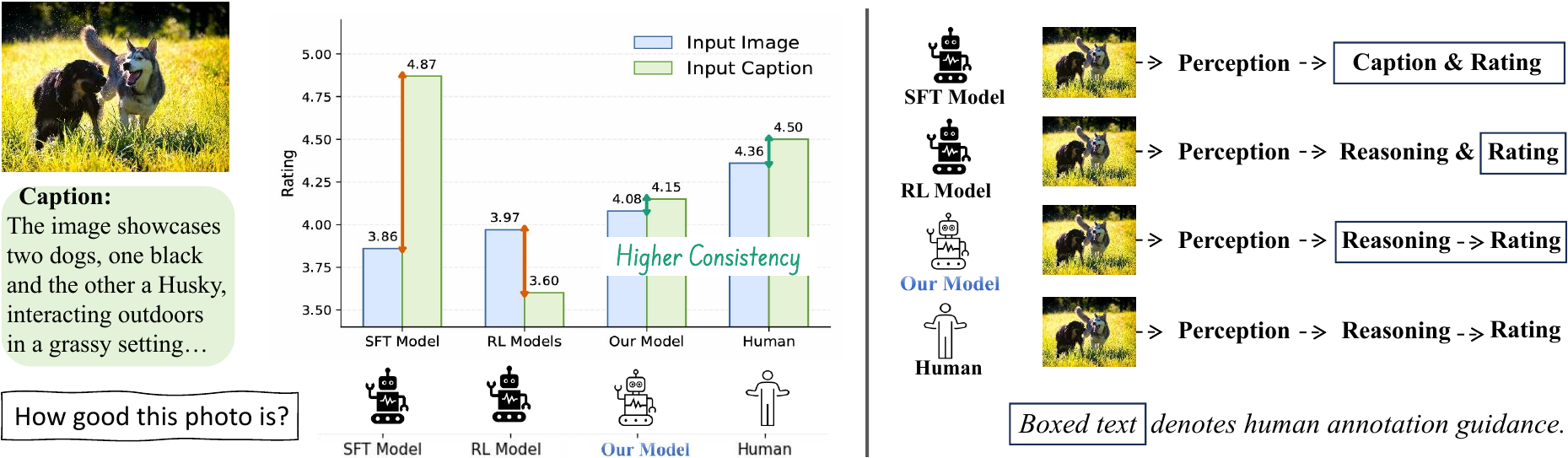}
    \vspace{-10pt}
    \captionof{figure}{\textbf{Aligning Model Reasoning with Human Judgments in Blind Image Quality Assessment.} \textbf{Left:} Comparison between image-conditioned and caption-conditioned quality evaluations. 
    Conventional models (here we test Q-Instruct model \cite{q-instruct} as the supervised fine-tuning (SFT) model and Q-Insight model \cite{q-insight} as the reinforcement learning (RL) model) yield inconsistent scores between image and caption input, while our model aligns with human judgments with consistent scores between them. 
    \textbf{Right:} Illustration of quality reasoning processes across model types.
    SFT-based models are supervised on captions and ratings but lack explicit reasoning guidance; existing RL-based models focus on score optimization.
     Humans reason about image quality through interpretable judgment criteria, enabling consistent assessment with or without direct visual input.
    Our model is jointly guided on reasoning and rating, mirroring the human evaluation process.
   } 
    \label{fig_1}
\end{center}%
}]

\vspace{3mm}

\maketitle

\begin{abstract}
Humans assess image quality through a perception-reasoning cascade, integrating sensory cues with implicit reasoning to form self-consistent judgments. In this work, we investigate how a model can acquire both human-like and self-consistent reasoning capability for blind image quality assessment (BIQA).
We first collect human evaluation data that capture several aspects of human perception-reasoning pipeline.
Then, we adopt reinforcement learning, using human annotations as reward signals to guide the model toward human-like perception and reasoning.
To enable the model to internalize self-consistent reasoning capability, we design a reward that drives the model to infer the image quality purely from self-generated descriptions.
Empirically, our approach achieves score prediction performance comparable to state-of-the-art BIQA systems under general metrics, including Pearson and Spearman correlation coefficients.
In addition to the rating score, we assess human–model alignment using ROUGE-1 to measure the similarity between model-generated and human perception–reasoning chains.
On over 1,000 human-annotated samples, our model reaches a ROUGE-1 score of 0.512 (cf. 0.443 for baseline), indicating substantial coverage of human explanations and marking a step toward human-like interpretable reasoning in BIQA.
\end{abstract}
\vspace{-15pt}    
\section{Introduction}
\label{sec:intro}
\vspace{-5pt}

Blind image quality assessment (BIQA) aims to simulate how humans perceive and evaluate the visual quality of an image.
To understand which visual features are extracted from the image in perception and how these features are logically integrated into an overall judgment, researchers have explored a wide range of computational approaches.

Early studies \cite{dbcnn,musiq,arniqa,topiq} achieved some level of  success by focusing on low- and mid-level visual features.
Using contrastive learning or visual feature encoders, these methods could effectively predict numerical quality scores.
However, as multimodal models \cite{llava,flamingo,mplug2,qwen2.5,yu2025learning} evolve, the research community has begun to seek interpretability beyond score regression—expecting models not only to rate image quality but also to articulate why an image appears better or worse.

This shift has inspired a new wave of pioneering work that bridges vision and language.
Recent studies \cite{liqe,depictqa,q-instruct,q-align,q-ground,q-insight,q-ponder} have explored semantic-based classification, textual explanation generation, and multimodal regression guided by visual features.
These models laid the foundation for multimodal BIQA, demonstrating the potential of combining image and text representations.
Nevertheless, they still fall short of replicating the human process of perception and reasoning. Rather than integrating perceptual cues into a logical judgment, they often directly generate both explanations and scores from image embeddings, so the two are often not logically connected.
As a result, these models may appear to reason, yet their “reasoning” remains shallow and directly coupled with visual input.

 As illustrated in Fig. \ref{fig_1}, existing multimodal large language model (MLLM) -based BIQA models can be roughly divided into two categories.
The first category is supervised fine-tuning (SFT) models, which lack a genuine reasoning process and treat image captions merely as by-products of rating.
The second category is reinforcement learning (RL) models, where the reasoning process is generated jointly with the quality score, but without human supervision.
To build a system that more closely resembles humans,
we propose to supervise the perception-reasoning stage by human annotations in an RL framework.
Before detailing the method, we first formalize the concept of human perception-reasoning in BIQA:
(1) Perception – visual image is transformed into internal representations, including low-level visual features and high-level semantic features;
(2) Reasoning – these representations are integrated into a coherent quality judgment.
By explicitly modeling this two-stage process, we enable the system to analyze intermediate textual information, simulate human perceptual focus, and reconstruct human evaluation criteria—thereby enhancing interpretability.

Our contributions are fourfold:
\begin{itemize}
    \item We collect human annotations on 1,495 images spanning eight dimensions related to image quality, which directly capture the human perception and reasoning process.
    \item We design new reward functions that enable the model to effectively evolve toward self-consistent and human-like reasoning under these human-annotated signals.
    \item We introduce ROUGE-1 as a metric for evaluating the alignment between model-generated and human perception–reasoning chains, providing a new direction for measuring human-aligned reasoning in BIQA.
    \item Our model achieves competitive performance under both image-based and caption-based conditions, offering a step toward interpretable, human-aligned BIQA.
\end{itemize}

\section{Related Works}

\subsection{Image Quality Assessment Datasets}
A variety of publicly available datasets \cite{csiq,spaq,koniq,kadid,live-w,agiqa,q-instruct,q-ground} have greatly advanced research in image quality assessment (IQA).
These datasets include large collections of both real-world and synthetically degraded images, significantly broadening the scope of quality evaluation studies.
From the perspective of annotation, in addition to the mandatory mean opinion score (MOS), some datasets provide supervisory signals.
For example, KonIQ~\cite{koniq} supplements MOS with score distributions and feature-level attributes such as contrast, colorfulness, and sharpness, while Q-Pathway \cite{q-instruct} introduces rich textual annotations that describe low-level perceptual attributes and visual content.
We categorize these as perceptual features, which reflect how humans perceive visual quality through sensory cues.

However, image features are inherently complex, and the same texture pattern may influence perceived quality differently across contexts.
What remains largely missing are authentic annotations that capture reasoning-level cues—how humans interpret and integrate perceptual information to form consistent judgments of quality.
To address this gap, we collect human evaluation data that explicitly describe how specific perceptual features affect perceived quality, providing a foundation for modeling the reasoning process underlying human IQA.
\subsection{From Score to Textual Reasoning for BIQA}
Early researchers primarily focused on predicting a numerical quality score.
Most existing methods \cite{constrastive-iqa,arniqa,topiq,multiscale} adopt a visual encoder–regression framework, where an image is embedded into a high-dimensional feature space and then mapped to a quality score through a regression network.
Various strategies have been explored under this paradigm, such as contrastive learning \cite{moco,contrastive} for different degradation types and multi-scale feature aggregation \cite{topiq,multiscale} across visual layers.
Although these models achieve strong performance, their interpretability remains limited, as they provide little insight into why certain images are judged as high or low quality.

With the rapid progress of multimodal reasoning and image–language understanding, recent studies have sought to enhance the interpretability of BIQA through natural language supervision.
Methods such as Q-Ponder\cite{q-ponder}, Q-Insight \cite{q-insight}, Q-Instruct \cite{q-instruct}, Co-Instruct \cite{co-instruct} and DepictQA \cite{depictqa} incorporate textual descriptions during quality prediction, aiming to capture human-like reasoning behind perceptual judgments.
These approaches mark a shift from purely perceptual modeling to language-guided quality assessment, offering new perspectives on how humans integrate perception and reasoning in evaluating image quality. Nevertheless, as illustrated in Fig. \ref{fig_1}, existing models either lack an explicit reasoning process or perform reasoning without effective supervision. 

\begin{figure*}[t]
    \centering
    \includegraphics[width=\linewidth]{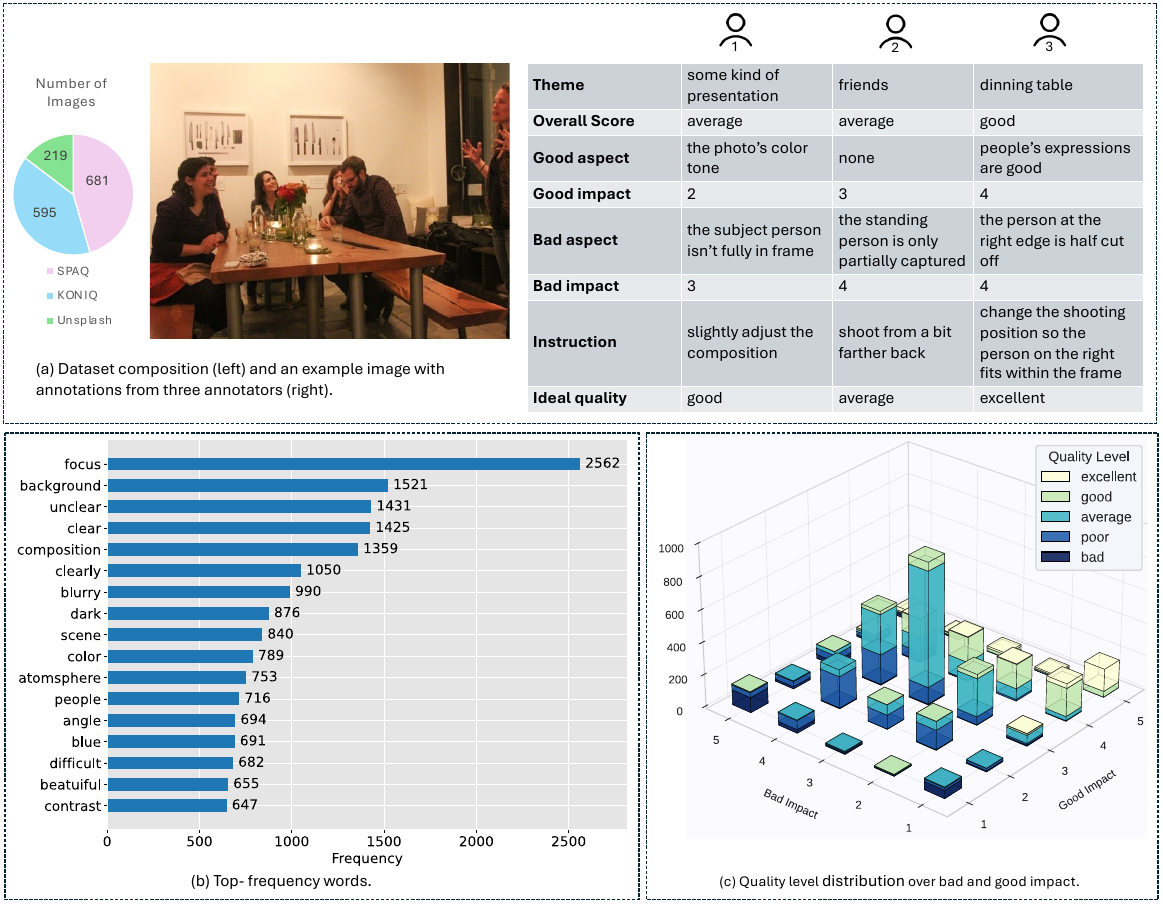}
    \caption{\textbf{Overview of the Q-Reasoning dataset.}
    }
    \label{fig_2}
\vspace{-10pt}
\end{figure*}
\vspace{-10pt}


\section{Methods}

\subsection{Q-Reasoning: Reasoning Annotation Dataset}
\label{4.1}
Previous works \cite{q-instruct, q-ground} have collected abundant low-level perceptual annotations for BIQA, such as texture, distortion, noise, and exposure bias.
However, these datasets lack the interpretive reasoning process that connects perceptual cues to final quality judgments.

To construct an interpretable reasoning pathway, we collect a human reasoning annotation dataset, which we refer to as Q-Reasoning dataset. As illustrated in Fig.~ \ref{fig_2} (a), Q-Reasoning totally contains 1,495 image samples and each sample has at least three evaluators.
We collected human reasoning annotations through an online crowdsourcing interface.
Participants first selected their native language and completed a brief tutorial with six example images.
They then answered the structured questions in their own language.
To ensure annotation quality and reduce fatigue, each annotator was limited to 15 images.
In total, about 300 participants contributed approximately 4,500 annotated samples.
Our image data is partially sampled from three existing datasets: two authentic BIQA datasets, SPAQ \cite{spaq} and KonIQ \cite{koniq}, and the Unsplash Lite Dataset \cite{unsplash}.
 Q-Reasoning captures how humans evaluate image quality by annotating eight aspects, including,
(1) semantic theme,
(2) overall quality score,
(3) good impact,
(4) scale of good impact,
(5) bad impact,
(6) scale of bad impact,
(7) suggestions,
(8) ideal quality.

In Fig.~\ref{fig_2} (b) and (c), we present the word frequency statistics of human attention and the joint distribution between quality scores, good scales, and bad scales on the raw data. From Fig.~\ref{fig_2} (c), we observe that good and bad impact factors jointly shape human quality perception: images may still receive a high quality score despite certain defects, whereas the opposite—strong defects leading to high scores—is far less common.
Compared with existing datasets, our collected data provides a direct reasoning pathway for image quality assessment.
\begin{table}[h]
\centering
\caption{Annotation dimensions between IQA datasets. Here L-Reasoning and H-Reasoning denote reasoning based on low-level and high-level visual features. Q-Pathway \cite{q-instruct} mainly targets low-level perception, while Q-Ground \cite{q-ground} inherits this focus and lacks authentic high-level semantic reasoning labels.}
\label{tab_1}
\resizebox{\columnwidth}{!}{
\begin{tabular}{lccccc}
\toprule
\textbf{Dataset} & \textbf{MOS} & \textbf{Captions} & \textbf{L-Reasoning}& \textbf{H-Reasoning} & \textbf{Impact Scale} \\
\midrule
Traditional datasets & \checkmark &  \\
Q-Pathway \cite{q-insight} & \checkmark & \checkmark & \checkmark \\
Q-Ground \cite{q-ground} & \checkmark & \checkmark & \checkmark \\
Q-Reasoning (Ours) & \checkmark & \checkmark & \checkmark & \checkmark & \checkmark \\
\bottomrule
\end{tabular}
}
\end{table}

Furthermore, we compute the Pearson correlation coefficient (PLCC) and Spearman correlation coefficient (SRCC) between the good/bad scales and the overall/ideal quality scores to validate our data reliability, as shown in Table~\ref{tab_2}.

\begin{table}[h]
\centering
\caption{PLCC / SRCC between human-annotated good/bad scales and overall / ideal quality scores.}
\label{tab_2}
\begin{tabular}{lrr}
\toprule
\textbf{} & Overall Quality & Ideal Quality \\
\midrule
Good Scale & 0.826 / 0.744 & 0.878 / 0.789 \\
Bad Scale  & 0.216 / 0.073 & 0.549 / 0.392 \\
\bottomrule
\end{tabular}
\end{table}

\begin{figure*}
    \centering
    \vspace{-2.3em}
    \includegraphics[width=\linewidth]{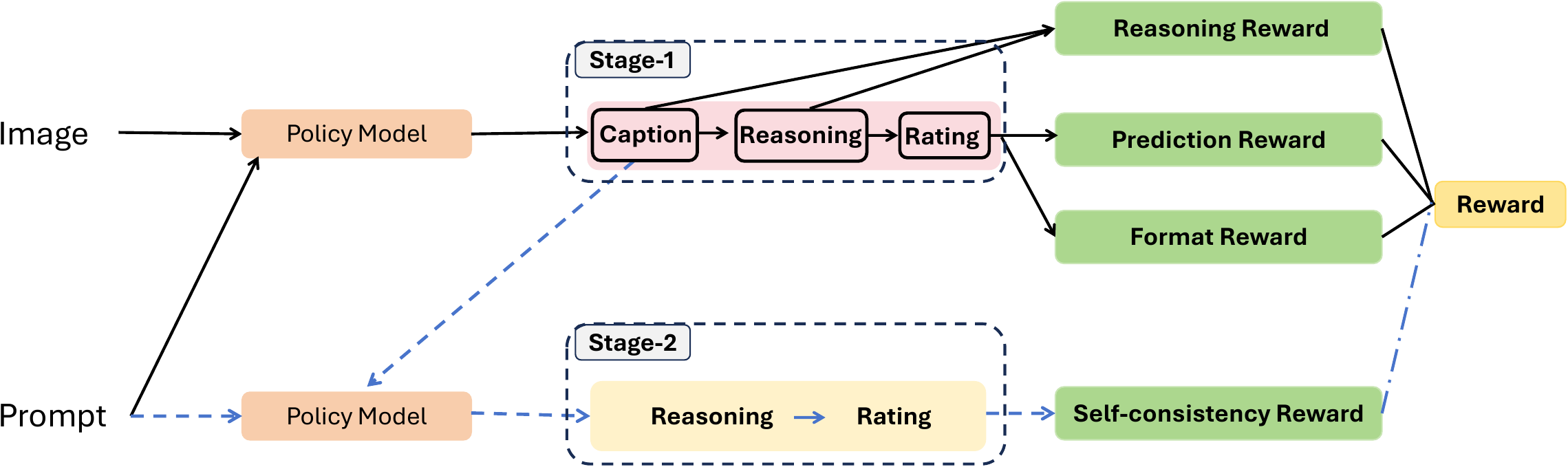}
    \caption{\textbf{Overview of the Proposed Human-Like Reasoning Framework.} The training process involves two reasoning stages.
    In the \textit{first reasoning stage}, the model receives both an image input and a textual prompt.
    The total reward combines three components: 
    (1) \textit{Reasoning reward}, measuring the similarity between the model’s generated explanation and human annotations;
    (2) \textit{Prediction reward}, aligning the predicted score with human ratings; and 
    (3) \textit{Format reward}, enforcing structural consistency in the output.
    In the \textit{second reasoning stage}, the model takes its previously generated caption and the same prompt as input, 
    and is optimized with a \textit{Self-consistency reward}.
    This dual-stage design encourages the first-stage policy to learn human-like perception and quality judgment,
    while the second-stage reasoning promotes internalization of human-like \textit{judgment criteria}.}
    \label{fig_3}
\vspace{-10pt}
\end{figure*}

\subsection{Overview of the Proposed Framework}
Our goal is to enable an MLLM to assess image quality through a human-like and self-consistent perception–reasoning process.
As illustrated in Fig.~\ref{fig_3}, the proposed framework follows a two-stage learning paradigm that integrates human-guided reinforcement and self-consistent reasoning. This design allows the model to align its perceptual and reasoning behavior with human judgments while maintaining self-consistent reasoning capability under both image-based and caption-based conditions.

In the first stage, the model learns human-consistent perception and reasoning through a reasoning reward guided by human annotations. In the second stage, we introduce a self-consistent reasoning objective that enables the model to infer image quality from its own textual captions. These two stages form a unified human-aligned BIQA framework that disentangles perception from reasoning.
\vspace{-3pt}
\subsection{Human-Guided Perception and Reasoning}
To guide the model toward human-like perception and reasoning while avoiding the template-style outputs often caused by SFT,
we introduce a relaxed guidance reward, denoted as the Reasoning Reward in Fig.~\ref{fig_3}.
During the first stage, the model generates three components:
\texttt{<caption>}, \texttt{<reasoning>}, and \texttt{<rating>}.
The reasoning reward measures how closely the model’s \texttt{<caption>} and \texttt{<reasoning>} align with human annotations, assigning a continuous reward value within the range of [0, 1].
Specifically, we use the ROUGE-1 \cite{rouge} score to quantify the similarity between model outputs and human annotated perception and reasoning. To simplify notation, we use \textit{human} to represent the human annotations 
and \textit{model} to represent the model-generated \texttt{<caption>} and \texttt{<reasoning>} components 
in the following formulations.
\begin{align}
&r_{\text{reasoning}} = \text{ROUGE-1}(model,human)\\
&\text{ROUGE-1} = 
\frac{\sum_{w \in V} \min \big( C_{human}(w),\, C_{model}(w) \big)}
{\sum_{w \in V} C_{human}(w)} 
\end{align}
where $V$ denotes the set of all unigrams appearing in the human annotations, 
$C_{human}(w)$ and $C_{model}(w)$ represent the occurrence counts of word $w$ 
in the human reference and the model-generated text, respectively. In essence, ROUGE-1 \cite{rouge} evaluates how much the model’s generated text overlaps with human annotations---a perfect match yields a reward of~1, otherwise, results in a score approaching~0.

\subsection{Self-Consistent Reasoning via Language}
In text-generative BIQA systems, previous works lack effective evaluation methods to assess the textual quality and reasoning logic of the generated explanations.
How to evaluate whether a model’s textual description of an image is semantically accurate and whether its reasoning is logically sound remains an open problem.

We propose an equivalent evaluation principle:
if a model can accurately predict image quality solely from its own generated caption, it has acquired a self-consistent reasoning mechanism that approximates human reasoning.
Following this principle, we introduce a text-only self-consistency reward that explicitly reinforces the model’s internal reasoning alignment.
Specifically, as illustrated in Fig.~\ref{fig_3}, we feed the model’s self-generated \texttt{<caption>} back as input and require it to infer the corresponding image quality without visable image. To encourage prediction consistency while maintaining smooth reward decay, 
we design the  self-consistency reward via a cosine-based function defined as:
\begin{equation}
r_{\text{self-consistency}} =
\begin{cases}
0.5 \big( 1 + \cos(\pi x / t) \big), & \text{if } x < t, \\
0, & \text{otherwise},
\end{cases}
\label{eq:score_reward}
\end{equation}
where $x = |s_{\text{pred}} - s_{\text{gt}}|$ denotes the absolute difference 
between the predicted and ground-truth quality scores, $t \in [0, 1]$ is a threshold hyperparameter controlling the reward bandwidth. 
As $x \to 0$, the reward approaches~1, indicating high alignment between prediction and ground truth, 
while $x \to t$ leads to $r_{\text{reasoning}} \to 0$, penalizing large deviations.

\subsection{Overall Training via GRPO Strategy}
We have introduced the designs of the reasoning and self-consistentcy rewards. 
In addition, two commonly used rewards in Group Relative Policy Optimization (GRPO~\cite{guo2025deepseek}) and BIQA frameworks~\cite{q-insight,q-ponder} are also adopted in our setup. 
The first is the \textit{image-based score prediction reward}, which evaluates the model’s numerical accuracy. 
In our design, the prediction reward is computed using the same smooth cosine formulation as Eq.~(\ref{eq:score_reward}). 
The second is the \textit{format reward}, defined as:
\begin{equation}
r_{\text{format}} =
\begin{cases}
0.5, & \text{if the output format is correct},\\
0, & \text{otherwise}.
\end{cases}
\label{eq:format_reward}
\end{equation}

Finally, the overall reward is obtained as a weighted combination of all components:

\begin{equation}
\begin{aligned}
r_{\text{total}} =\;& 
\underbrace{r_{\text{reasoning}}}_{\text{Human-guided}} \\
& + \underbrace{r_{\text{self-consistency}} + r_{\text{prediction}}}_{\text{Score-guided}}
+ r_{\text{format}} .
\label{eq:total_reward}
\end{aligned}
\end{equation}

After computing all reward terms, we integrate them into the GRPO optimization framework.
The GRPO objective is defined as follows:

\begin{equation}
\begin{aligned}
    \mathcal{J} _{GRPO} =  \mathbb{E}[\frac{1}{N}{\textstyle \sum_{i}^{N}}min(d_iA_i, C_{d_i,\epsilon}A_i-\beta\cdot \text{KL})],
\end{aligned}
\label{grpo}
\end{equation}
where $d_i = \frac{\pi_{\theta}(y_i|x)}{\pi_{{\theta}_{\text{old}}}(y_i|x)}$, $A_i = \frac{r_i-mean(r_1,r_2,...,r_N)}{std(r_1,r_2,...,r_N)}$, $C_{d_i,\epsilon}=clip(d_i,1-\epsilon,1+\epsilon)$, and $\text{KL} = \mathbb{D}_{\text{KL}}(\pi_{\theta}||\pi_{\text{ref}})$. Note that $\pi_{\theta}$, $\pi_\text{old}$ and $\pi_\text{ref}$ denote the policy model, old policy model and reference model, respectively. $r_i$ denotes rewards, and $\epsilon$ and $\beta$ denote hyper-parameters.

Now, let us refer to Eq.~(\ref{grpo}) with a specific reasoning stage. For an arbitrary image, the model generates $N$ reasoning trajectories $\{r_i\}_{i=1}^N$, each assigned with a scalar reward $R_i$. 
After normalization within the group, we obtain the relative reward
$A_i^{(g)} = R_i - \bar{R}^{(g)}$. 
A positive $A_i^{(g)}$ indicates that the reasoning trajectory $r_i$ 
achieves higher quality than the group average—reflecting stronger alignment 
with our human-guided and score-guided joint objectives. 
During optimization, such trajectories are reinforced by increasing their 
generation probability, while those with negative advantages are suppressed. 
Through iterative updates, the model evolves towards both human-aligned reasoning and more accurate score estimation.
\section{Experiments}
\subsection{Experimental Settings}
\label{sec4.1}

\begin{table*}[t]
\renewcommand{\arraystretch}{1.15}
\centering
\caption{
\textbf{PLCC / SRCC performance comparison} between our model and other BIQA methods. 
All models (except for early hand-crafted ones) are trained on the KonIQ~\cite{koniq} training set. 
The best and second-best results are highlighted in \best{bold red} and \secondbest{underlined blue}.}
\label{tab:score}
\vspace{-5pt}

\resizebox{\textwidth}{!}{
\begin{tabular}{l|rrrrrrr}
\toprule
\textbf{Model} & \textbf{KonIQ~\cite{koniq}} & \textbf{SPAQ~\cite{spaq}} & \textbf{KADID~\cite{kadid}} & \textbf{LIVE-W~\cite{live-w}} & \textbf{AGIQA~\cite{agiqa}} & \textbf{CSIQ~\cite{csiq}} & \textbf{AVG.} \\
\midrule

\multicolumn{8}{@{\hskip 0pt}c@{\hskip 0pt}}{\textbf{\textit{Hand-Crafted Models}}} \\
\midrule
NIQE~\cite{NIQE} (2012) & 
0.533 / 0.530 & 0.679 / 0.664 & 0.468 / 0.405 & 0.493 / 0.449 & 0.560 / 0.533 & 0.718 / 0.628 & 0.575 / 0.535 \\
BRISQUE~\cite{BRISQUE} (2012) & 
0.225 / 0.226 & 0.490 / 0.406 & 0.429 / 0.356 & 0.361 / 0.313 & 0.541 / 0.497 & 0.740 / 0.556 & 0.464 / 0.392 \\

\midrule
\multicolumn{8}{@{\hskip 0pt}c@{\hskip 0pt}}{\textbf{\textit{Deep-Learning Models}}} \\
\midrule
NIMA~\cite{nima} (2018) & 
0.896 / 0.859 & 0.838 / 0.856 & 0.532 / 0.535 & 0.814 / 0.771 & 0.715 / 0.654 & 0.695 / 0.649 & 0.748 / 0.721 \\
DBCNN~\cite{dbcnn} (2019) & 
0.884 / 0.875 & 0.812 / 0.806 & 0.497 / 0.484 & 0.773 / 0.730 & 0.641 / 0.648 & 0.586 / 0.572 & 0.714 / 0.689 \\

MUSIQ~\cite{musiq} (2021) & 
0.924 / 0.929 & 0.868 / 0.863 & 0.575 / 0.556 & 0.789 / 0.830 & 0.722 / 0.630 & 0.771 / 0.710 & 0.775 / 0.753 \\
MANIQA~\cite{yang2022maniqa} (2022) & 
0.849 / 0.834 & 0.768 / 0.758 & 0.499 / 0.465 & 0.849 / 0.832 & 0.723 / 0.636 & 0.623 / 0.627 & 0.719 / 0.692 \\
CLIP-IQA+~\cite{wang2023exploring} (2023) & 
0.909 / 0.895 & 0.866 / 0.864 & 0.653 / 0.654 & 0.832 / 0.805 & 0.736 / 0.685 & 0.772 / 0.719 & 0.795 / 0.770 \\

\midrule
\multicolumn{8}{@{\hskip 0pt}c@{\hskip 0pt}}{\textbf{\textit{SFT-based and RL-based MLLMs}}} \\
\midrule
C2Score~\cite{zhu2024adaptive} (2024) &
0.923 / 0.910 & 0.867 / 0.860 & 0.500 / 0.453 & 0.786 / 0.772 & 0.777 / 0.671 & 0.735 / 0.705 & 0.765 / 0.729 \\
Q-Align~\cite{q-align} (2024) &
\secondbest{0.941 / 0.940} & 0.886 / 0.887 & 0.674 / 0.684 & 0.853 / 0.860 & 0.772 / 0.735 & 0.671 / 0.737 & 0.799 / 0.807 \\
DeQA~\cite{deqa} (2025) &
\best{0.953 / 0.941} & \best{0.895} / 0.896 & \secondbest{0.694} / 0.687 & \best{0.892 / 0.879} & \secondbest{0.809} / 0.729 & \secondbest{0.787 / 0.744} & \best{0.838} / \secondbest{0.813} \\

Q-Insight-Score~\cite{q-insight} (2025) & 
0.918 / 0.895 & 0.887 / \secondbest{0.899} & 
\best{0.702} / \secondbest{0.702} & 0.870 / 0.839 &  \best{0.816 / 0.766} &
0.685 / 0.640 &  0.813 / 0.789\\

\textbf{Ours} &
0.930 / 0.920 & \secondbest{0.893} / \best{0.907} & 0.672 / \best{0.734} & \secondbest{0.877 / 0.849} & 0.803 / \secondbest{0.760} & \best{0.842 / 0.823} & \secondbest{0.836} / \best{0.832} \\
\midrule

\multicolumn{8}{@{\hskip 0pt}c@{\hskip 0pt}}{\textbf{\textit{Caption-Only Conditions}}} \\
\midrule

Q-Insight-Score~\cite{q-insight} (2025) & 
0.841 / 0.818 & 0.840 / 0.847 & 
0.645 / 0.650 & 0.769 / 0.757 &  
0.639 / 0.592 & 0.752 / 0.677 & 0.748 / 0.724  \\
\textbf{Ours} &
\textbf{0.871 / 0.855} & \textbf{0.861 / 0.875} & \textbf{0.662 / 0.676} & \textbf{0.791 / 0.798} & \textbf{0.783 / 0.707} & \textbf{0.788 / 0.727} & \textbf{0.812 / 0.772} \\
\bottomrule
\end{tabular}
}
\vspace{-8pt}
\end{table*}

\textbf{Training Dataset} To ensure a fair comparison with previous BIQA methods, we train our model on a subset of the KonIQ \cite{koniq} dataset containing 7{,}058 images, 
following the same data configuration used in prior works \cite{deqa,q-insight}. 
It is worth noting that our collected \textit{Q-Reasoning} dataset 
shares 482 images with this subset. 
For these overlapping images, the corresponding human annotations 
from \textit{Q-Reasoning} are utilized during training to provide additional supervision signals for perception and reasoning alignment.\\
\textbf{Evaluation Datasets} \label{4.1.2} For evaluation, we employ a total of seven datasets, including six existing benchmarks and the collected Q-Reasoning dataset in this work.
The authentic image quality datasets consist of SPAQ \cite{spaq}, KonIQ\cite{koniq}, LIVE-W\cite{live-w}, and our collected dataset,
while the synthetic datasets include CSIQ \cite{csiq}, AGIQA \cite{agiqa}, and KADID~\cite{kadid}.
Although our collected dataset partially overlaps with SPAQ \cite{spaq} and KonIQ \cite{koniq} in image content, the training and testing splits are strictly separated, ensuring that no image is shared across phases.\\
\textbf{Model Parameter Settings} We adopt Qwen2.5-VL-7B-Instruct \cite{qwen2.5} as the MLLM backbone. We warm-up the models as the settings in \cite{q-insight}, which we accordingly adopt as the baseline in our experiments. We adopt LoRA-based fine-tuning with rank = 8 and $\alpha = 16$.
During reinforcement learning, we apply the GRPO algorithm with a group number of 4, 
while keeping other hyperparameters at their default settings. We employ the AdamW~\cite{adam} optimizer with an initial learning rate of $1\times10^{-6}$. The batch size is set to 2, and the model is trained for two epochs on eight NVIDIA A6000 GPUs. Training on the KonIQ \cite{koniq} subset for 2 epochs takes approximately 44 hours to complete.

\begin{figure*}[h!]
    \centering
    \includegraphics[width=1.0\textwidth]{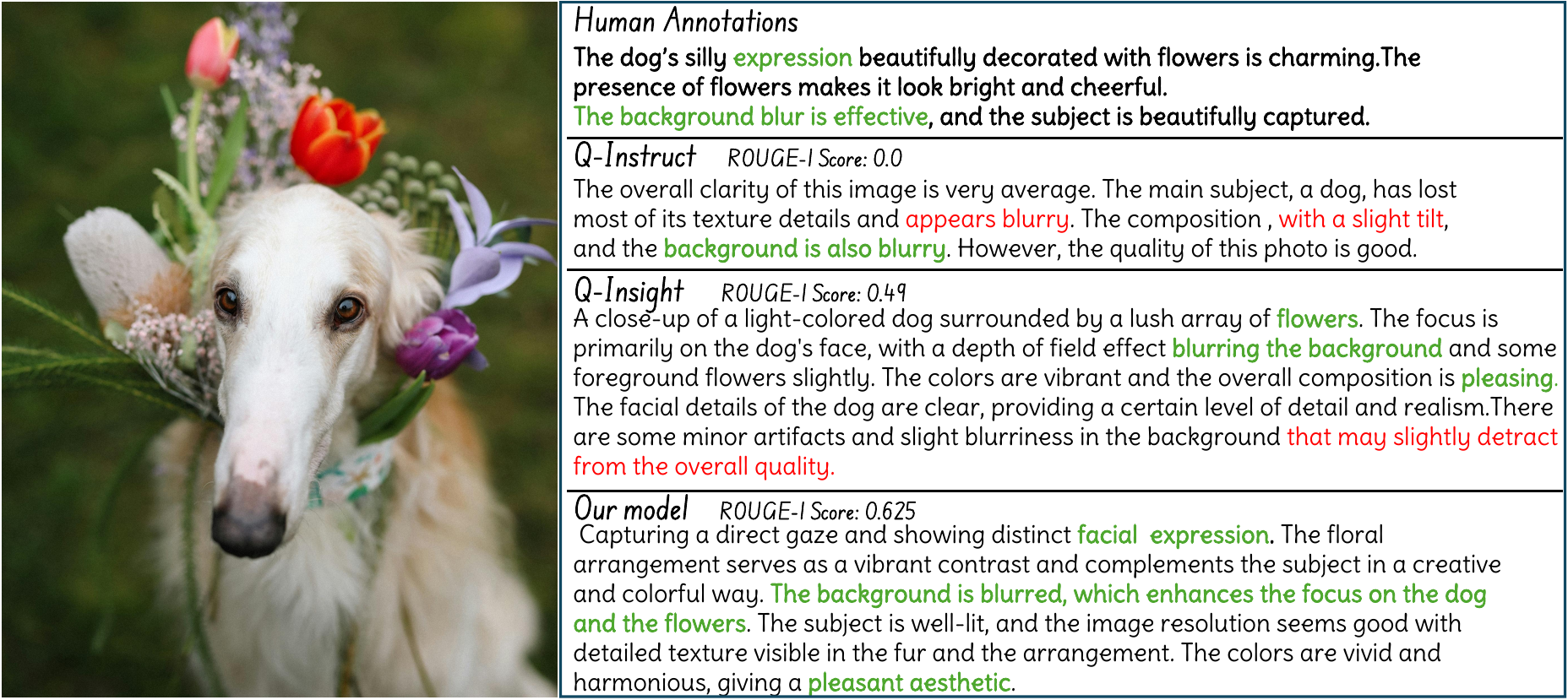}
    \caption{\textbf{Case study on model-human reasoning alignment.}
We compare the SFT-based \textbf{Q-Instruct}~\cite{q-instruct}, the RL-based \textbf{Q-Insight-Score}~\cite{q-insight}, and \textbf{our model}.
\textcolor[rgb]{0,0.6,0}{Green text} indicates reasoning parts consistent with human annotations,
while \textcolor[rgb]{0.8,0,0}{red text} highlights inconsistencies.
ROUGE-1 score \cite{rouge} measures how well the model's reasoning captures human reasoning content.}
    \label{fig_case_study}
\vspace{-15pt}
\end{figure*}

\subsection{Image-based Score Comparison}
To evaluate the model’s performance under image-conditioned settings, we compare our model with three major categories of BIQA methods: traditional hand-crafted algorithms (NIQE~\cite{NIQE} and BRISQUE~\cite{BRISQUE}), deep-learning–based regression models (NIMA~\cite{nima}, DBCNN~\cite{dbcnn}, MUSIQ~\cite{musiq}, MANIQA~\cite{yang2022maniqa}, CLIP-IQA+~\cite{wang2023exploring}), and recent multimodal models trained with SFT (C2Score~\cite{zhu2024adaptive}, Q-Align \cite{q-align} and DeQA~\cite{deqa}) or RL (Q-Insight~\cite{q-insight}).
For fair comparison, all models should be trained on the same split of KonIQ \cite{koniq} dataset. However, since Q-Insight \cite{q-insight} adopts a joint-training strategy that includes an additional 7K images beyond the standard training split, we report the performance of Q-Insight-Score \cite{q-insight} to represent this baseline fairly.
We conduct evaluations on six widely used datasets mentioned in Sec.~\ref{4.1.2}. Experimental results demonstrate that our model achieves competitive performance with state-of-the-art methods across all the datasets. We also observe that SFT-based methods achieve relatively high performance under image-conditioned evaluation.
However, this comes at the cost of their textual generation and reasoning capabilities, which are often weakened due to overfitting to score supervision during fine-tuning. Compared with the RL-based baseline Q-Insight-Score \cite{q-insight} model, our approach achieves an average improvement of 0.023 PLCC and 0.041 SRCC across test sets. Nevertheless, our model exhibits weaker performance than the baseline on a few datasets, and we analyze possible underlying reasons in Sec.~\ref{discussion}.

\subsection{Caption-based Score Comparison}
We further evaluate the model’s self-consistency by testing whether it can still assess image quality even when visual input is removed after the caption is generated.
Unfortunately, we find that most SFT-based MLLMs fail to process text-only inputs.
Therefore, we compare the performance of Q-Insight-Score\cite{q-insight} and our model across different datasets under the caption-only condition, as shown at the bottom of Table~\ref{tab:score}.
To eliminate potential answer cues, we explicitly instruct the model to describe only visible content in its captions.
Our model demonstrates strong self-consistency and generalization capabilities, enabling it to estimate image quality solely from textual captions.

\subsection{Human Consistency Evaluation}
\vspace{-5pt}
During training, we employed ROUGE-1~\cite{rouge} to measure the consistency between the model’s process and human. Here, we adopt the same metric to evaluate our model and the baseline Q-Insight-Score~\cite{q-insight} on the Q-Reasoning subset, consisting of 1,013 human-annotated samples.
Q-Insight-Score~\cite{q-insight} gains a 0.443 for ROUGE-1 score~\cite{rouge}. Our model further improves the alignment with human reasoning by approximately 0.07 in ROUGE-1 \cite{rouge} score, finally gains 0.512. We then conducted a qualitative case study, as illustrated in Fig.~\ref{fig_case_study}.
We compare the Q-Instruct~\cite{q-instruct}, Q-Insight-Score~\cite{q-insight}, and our model.
The results show that while all models exhibit similar accuracy in low-level perceptual descriptions, only our model produces human-like expressions and demonstrates an understanding of aesthetic intent—for example, recognizing that background blurring enhances the prominence of the main subject.
This behavior suggests that our model better internalizes the reasoning patterns humans employ in visual quality assessment.

Although ROUGE-1~\cite{rouge} is effective for evaluating textual overlap and summarization quality, it fails to capture semantic discrepancies. For instance, Q-Instruct once described the dog as blurred, which was factually incorrect, yet still received a ROUGE-1\cite{rouge} reward because the human annotation contained the word “blur”. This highlights the need for reasoning-aware metrics that align more faithfully with human semantic understanding.

\begin{table*}[t]
\renewcommand{\arraystretch}{1.15}
\centering
\caption{
\textbf{Ablation study on reasoning and  self-consistency rewards.}
We evaluate the contribution of each reward term under both image-only and caption-only conditions.
Results are reported as PLCC / SRCC (↑ higher is better).
Training and inference times are measured on the KonIQ~\cite{koniq} dataset.
}
\label{tab:ablation}
\vspace{-6pt}

\resizebox{\textwidth}{!}{
\begin{tabular}{l|rrrrrr|r|r|rr}
\toprule
\textbf{Setting} & KonIQ & SPAQ & KADID & LIVE-W & AGIQA & CSIQ & \textbf{AVG.} & Q-Reasoning & Train (hrs/epoch) & Infer (s/img) \\
\midrule

\multirow{2}{*}{Baseline} 
& 0.918 / 0.895 & 0.887 / 0.890 & \textbf{0.702} / 0.702 & 0.870 / 0.839 & 0.802 / 0.759 & 0.685 / 0.640 & 0.806 / 0.789 & \multirow{2}{*}{0.443} & \multirow{2}{*}{--}  & \multirow{2}{*}{5.95 / 3.60} \\
& \textit{0.841 / 0.818} & \textit{0.840 / 0.847} & \textit{0.645 / 0.650} & \textit{0.769 / 0.757} & \textit{0.639 / 0.592} & \textit{0.752 / 0.677} & \textit{0.748 / 0.724} &  &  \\

\midrule
\multirow{2}{*}{+ reasoning Reward} 
& 0.927 / 0.913 & 0.886 / 0.878 & 0.670 / 0.712 & 0.873 / 0.835 & 0.802 / 0.761 & 0.762 / 0.731 & 0.820 / 0.805 & \multirow{2}{*}{0.494}& \multirow{2}{*}{$\approx$11}& \multirow{2}{*}{5.68 / 3.32} \\
& \textit{0.846 / 0.823} & \textit{0.847 / 0.856} & \textit{0.648 / 0.655} & \textit{0.766 / 0.755} & \textit{0.642 / 0.644} & \textit{0.723 / 0.724} & \textit{0.745 / 0.743} &  &  \\

\midrule
\multirow{2}{*}{+  Self-consistency Reward } 
& 0.929 / 0.919 & 0.893 / 0.907 & 0.675 / \textbf{0.735} & \textbf{0.879 / 0.851} & \textbf{0.800 / 0.760} & 0.832 / 0.818 & 0.834 / 0.831 & \multirow{2}{*}{0.457} & \multirow{2}{*}{$\approx$21.5} & \multirow{2}{*}{5.59 / 3.10}\\
& \textit{0.855 / 0.832} & \textit{0.854 / 0.857} & \textit{ \textbf{0.662} / 0.675} & \textit{\textbf{0.794} / 0.775} & \textit{0.767 / 0.685} & \textit{0.771 / 0.708} & \textit{0.784 / 0.757} &  &  \\

\midrule
\multirow{2}{*}{Full Model} 
& \textbf{0.930 / 0.920} & \textbf{0.893 / 0.907} & 0.672 / 0.734 & 0.877 / 0.849 & \textbf{0.800 / 0.760} & \textbf{0.842 / 0.823} & \textbf{0.836 / 0.832} & \multirow{2}{*}{0.512} & \multirow{2}{*}{$\approx$22}  & \multirow{2}{*}{6.30 / 3.58}\\
& \textit{\textbf{0.871 / 0.855}} & \textit{\textbf{0.861 / 0.875}} & \textit{\textbf{0.662 / 0.676}} & \textit{
0.791 / \textbf{0.798}} & \textit{\textbf{0.775 / 0.699}} & \textit{\textbf{0.788 / 0.727}} & \textit{\textbf{0.791 / 0.771}} &  &  \\

\bottomrule
\end{tabular}
}
\vspace{-8pt}
\end{table*}

\subsection{Ablation Study}
To examine the contribution of each component in our reward design, we conduct a series of ablation experiments.
All models are trained under identical settings, using the same data and hyperparameter configurations described in Section~\ref{sec4.1}.
For evaluation, inference is performed on a single NVIDIA A6000 GPU with a batch size of 1.

\begin{table}[t]
\centering
\renewcommand{\arraystretch}{1.15}
\caption{
\textbf{Preliminary ablation on format and prediction rewards.}
Evaluated on 178 KonIQ~\cite{koniq} test images. The format reward is set to either 0.5 or 1.0, and we compare the default discrete prediction reward against our smooth cosine–based variant.
}
\label{tab:format_prediction_ablation}
\vspace{-4pt}

\begin{tabular}{llr}
\toprule
Format & Prediction & PLCC / SRCC \\
\midrule
0.5 & discrete & 0.896 / 0.895 \\
0.5 & smooth   & \textbf{0.900 / 0.899} \\
1.0 & discrete & 0.896 / 0.895 \\
1.0 & smooth   & 0.899 / 0.897 \\
\bottomrule
\end{tabular}
\vspace{-15pt}
\end{table}

As a preliminary analysis, we start by isolating the effects of the format reward and the prediction reward.
Specifically, we compare two variants of the format reward (a fixed value of 0.5 vs. 1.0) and two variants of the prediction reward (a discrete reward vs. our smooth cosine-based reward), in Table~\ref{tab:format_prediction_ablation}. Based on these results, we adopt the configuration with a format reward of 0.5 and the smooth cosine–based prediction reward as our baseline for all subsequent experiments.

Next, we conduct a systematic ablation study on the reasoning reward and the  self-consistency reward .
Our analysis focuses on four aspects:
(1) the model’s quality prediction performance across multiple datasets,
(2) the model's self-consistency
(3) its alignment with human reasoning, and
(4) the training and inference time cost associated with each configuration.
The results are summarized in Table~\ref{tab:ablation}. 
From the ablation results, we observe that the full training configuration achieves the most balanced and robust overall performance.
A closer examination of each component further reveals complementary effects:
the  self-consistency reward  directly improves the model’s ability to fit quality scores by strengthening its internal inference mechanism,
while the reasoning reward primarily encourages the model to align its perceptual reasoning with human annotations.
Together, these two rewards jointly enhance both score prediction accuracy and human–model reasoning consistency. As the  self-consistency reward  requires the model to perform an additional inference step during training, it inevitably increases training time.
However, the inference-time cost remains essentially unchanged, since the deployed model executes a single forward pass at test time.

\vspace{-10pt}
\section{Discussion and Limitations}
\label{discussion}
\vspace{-2pt}
\textbf{Does Human-Like Reasoning Help Rating?}
In this work, we aim to align the model’s perceptual focus and reasoning patterns with those of humans.
Our experiments confirmed that the model indeed became more human-like in both perception and reasoning.
However, the ablation studies revealed an interesting phenomenon: as the model became more aligned with human reasoning, its score-prediction accuracy did not necessarily improve—and in some cases even declined.
This observation suggests that the optimization directions for human-like reasoning and for numerical score prediction are not perfectly aligned.
Human reasoning emphasizes interpretability, semantic abstraction, and multi-factor judgment, whereas BIQA score prediction—especially in existing benchmarks—tends to reward correlation with dataset-specific statistical cues.
As a result, being more human-like does not automatically imply better score prediction, indicating an inherent gap between human visual cognition and dataset-driven numerical supervision. This divergence highlights an important challenge for future BIQA research:
bridging the difference between human-centered reasoning and metric-centered prediction, and designing learning objectives that reconcile these goals.
\textbf{Limitations and Future Work.}
Despite these encouraging results, several limitations remain.
First, the scale of our human-annotated dataset is still limited and does not fully capture the diversity and complexity of human reasoning.
Second, our evaluation of human--model consistency relies primarily on ROUGE-1 \cite{rouge}, which measures lexical overlap but fails to capture deeper semantic correctness, sometimes rewarding incorrect reasoning due to shared vocabulary.
Third, while our framework enhances both prediction accuracy and interpretability, the self-consistency reward introduces additional training overhead due to the dual-inference design, which diverges from the efficiency of human learning.
These limitations point to several promising directions for future work: expanding the scope of human reasoning annotations, developing semantically grounded evaluation metrics, and exploring more efficient or scalable alignment strategies.
Ultimately, we hope this line of work contributes to building MLLM systems capable of reasoning about images in a principled, interpretable, and authentically human-aligned manner.

\vspace{-5pt}
\section{Conclusion}
\vspace{-5pt}
In this work, we explored how to model a human-like and self-consistent BIQA system.
We collected a new set of human perception–reasoning annotations and used them to guide the model toward human-aligned visual understanding.
In parallel, we introduced a caption-based self-consistency objective that required the model to infer quality solely from its own generated descriptions, thereby strengthening its internal reasoning ability. When jointly optimized, they produced a model that not only achieved strong rating performance but also exhibited substantially improved interpretability and human-model alignment. In addition to the traditional image-based BIQA setting, our work introduced two new evaluation dimensions for the field. We introduced caption-based BIQA and a metric for evaluating model–human alignment, moving the task beyond score prediction toward interpretable, human-driven quality assessment.
\clearpage
{
    \small
    \bibliographystyle{ieeenat_fullname}
    \bibliography{main}
}
\clearpage
\setcounter{page}{1}
\maketitlesupplementary

\setcounter{section}{0}

\renewcommand{\thesection}{S\arabic{section}}

\setcounter{figure}{0}
\renewcommand{\thefigure}{S\arabic{figure}}

\setcounter{table}{0}
\renewcommand{\thetable}{S\arabic{table}}

\setcounter{equation}{0}
\renewcommand{\theequation}{S\arabic{equation}}

\section{Extended Comparisons of Human Consistency Evaluation}

\begin{table*}[!b]
\renewcommand{\arraystretch}{1.15}
\centering
\caption{
\textbf{Human Consistency Evaluation.} ROUGE-1 scores~\cite{rouge} between model-generated perception-reasoning and human annotations. 
Our model achieves the highest human alignment (0.514), outperforming both SFT-based models (Q-Instruct~\cite{q-instruct}, DepictQA~\cite{depictqa}) and RL-based models (Q-Insight~\cite{q-insight}). }
\label{tab:s1}
\vspace{-5pt}

\resizebox{\textwidth}{!}{
\begin{tabular}{l|rrrrrr}
\toprule
\textbf{Model} & \textbf{Q-Instruct~\cite{q-instruct}} & \textbf{DepictQA~\cite{depictqa}} & \textbf{Q-Insight-Score~\cite{q-insight}} & \textbf{Q-Insight~\cite{q-insight}} & \textbf{Ours (base)} & \textbf{Ours (detailed)}  \\
\midrule

ROUGE-1 Score~\cite{rouge}  & 
0.279 & 0.318 & 0.443 & 0.487 & 0.512& \textbf{0.514}  \\

\bottomrule
\end{tabular}}
\vspace{-8pt}
\end{table*}

In the experimental section, we evaluated the human-consistency magnitude between the baseline model and our proposed model using the ROUGE-1 metric~\cite{rouge}.
In Table~\ref{tab:s1}, we also include SFT-based models such as Q-Instruct~\cite{q-instruct} (trained with extensive human annotations) and DepictQA~\cite{depictqa} (trained with semi-human supervision), as well as the RL-based Q-Insight~\cite{q-insight} (full) model.
Our model achieves the highest overlap ratio with human annotations (0.514), while SFT-based models capture only around 30\% of the human perception–reasoning process.
As illustrated in Fig.~\ref{fig_2}, degradations such as “blur” are among the most frequently mentioned features in human annotations, explaining why Q-Insight~\cite{q-insight} (full) exhibits stronger alignment with human reasoning than Q-Insight-Score~\cite{q-insight}. This is because, compared with the Q-Insight-Score~\cite{q-insight} model, the Q-Insight~\cite{q-insight} (full) model is additionally trained on a degradation perception task using extra data.
In Table~\ref{tab:s1}, we also include a detailed variant of our model, which will be discussed in the following section.
\vspace{-5pt}
\section{Detailed Variant and Prompt Robustness}
As discussed in Fig.~\ref{fig_2}, the Q-Reasoning dataset contains fine-grained human annotations, including semantic themes, advantages, flaws, etc.
To maximally leverage these detailed annotations, an intuitive approach is to design a structured prompt template that explicitly guides the model to follow a human-like reasoning format during training. To see the effects of prompt templates on human-alignment performance and prompt robustness, we compare the four systems using one of the following prompt templates.

The baseline score model (a) adopts two sections: think and answer.
In contrast, our model (b) introduces three sections: caption, think, and answer.
Furthermore, detailed variants of our model, (c) and (d), expand the template by adding more components, including subject, flaw, and other reasoning-related fields.
\vspace{6pt}

\begin{tcolorbox}[title={(a) Baseline Prompt (Q-Insight-Score~\cite{q-insight})}]
\small
\texttt{" ... The reasoning process and answer are enclosed within <think> </think> and <answer> </answer> tags, respectively, i.e., <think> reasoning process here </think><answer> answer here </answer>"
}
\end{tcolorbox}

\begin{tcolorbox}[title={(b) Ours (Base Template)}]
\small
\texttt{" ... Follows a human thinking logics. ... The description, reasoning and answer are enclosed within <caption> </caption> <think> </think> and <answer> </answer> tags, respectively, i.e., <caption> description here </caption> <think> reasoning process here </think> <answer> answer here </answer>."
}
\end{tcolorbox}

\begin{tcolorbox}[title={(c) Ours ( Detailed Variant Template)}]
\small
\texttt{"... Follows a human thinking logics. ... i.e.,\\
<subject> the main subject </subject>\\
<advantage> the advantage  </advantage>\\
<flaw> the flaw </flaw>\\
<think> reasoning process </think>\\
<answer> answer </answer>."
}
\end{tcolorbox}

\begin{figure*}[!t]
    \centering
    \includegraphics[width=1.0\textwidth]{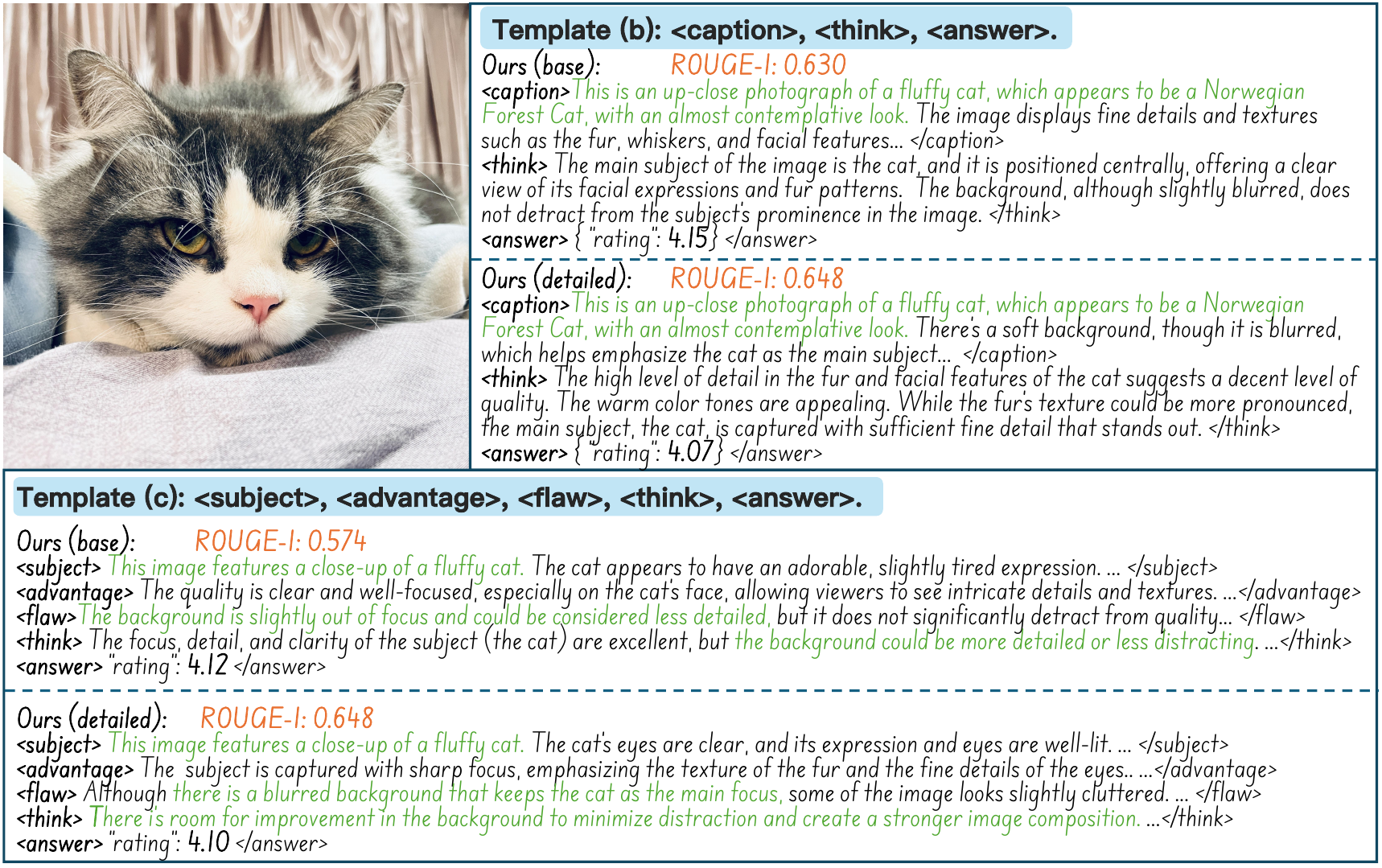}
    \caption{\textbf{Prompt Template Robustness.}
We evaluate our \textcolor{mblue}{base} and \textcolor{mblue}{detailed} models under 
template (b) and template (c). 
\textcolor[rgb]{0,0.6,0}{Green} highlights indicate perception-reasoning components that remain \textcolor{green!60!black}{consistent across templates}. 
Both models successfully adapt to new template patterns and produce stable quality predictions. 
The \textcolor{mblue}{detailed} model further demonstrates strong human alignment, achieving a 
ROUGE-1~\cite{rouge} score of \textbf{0.648} under both templates, indicating that its reasoning behavior remains 
consistent and robust regardless of the prompt structure.
}
    \label{fig:s1}
\vspace{-15pt}
\end{figure*}

\begin{tcolorbox}[title={(d) Ours ( Detailed Variant Template Version 2)}]
\small
\texttt{"... Follows a human thinking logics. ... i.e.,\\
<subject> the main subject </subject>\\
<advantage> the advantage  </advantage>\\
<flaw> the flaw </flaw>\\
<ideal> ideal quality </ideal>\\
<gap> gap between this one and ideal quality </gap>\\
<think> reasoning process </think>\\
<answer> answer </answer>."
}
\end{tcolorbox}
\vspace{4pt}
During training, template (d) frequently prevented the model from obtaining valid rewards for updates.
Therefore, we focus our analysis on templates (b) and (c) to examine their impact on prompt robustness. 
The model trained under template (b) is referred to as the \textcolor{mblue}{base} model, 
while the one trained under template (c) is referred to as the \textcolor{mblue}{detailed} model. 
As shown in Fig.~\ref{fig:s1}, we evaluate both models using template (b) and (c), respectively, 
to assess their generalization and robustness to prompt variations. Both the base and detailed models successfully adapt to unseen templates (e.g., template (b) for the detailed model and template (c) for base model), demonstrating strong prompt and format robustness.
Interestingly, both models produce similar contents under the first template, suggesting that during training, the model focuses more on learning the underlying reasoning knowledge rather than overfitting to a specific prompt structure—consistent with observations reported in DeepSeek-R1~\cite{guo2025deepseek}.

\vspace{-5pt}
\section{Detailed Case Studies on Perception-Reasoning Levels}
\begin{figure*}[!h]
    \centering
    \includegraphics[width=1.0\textwidth]{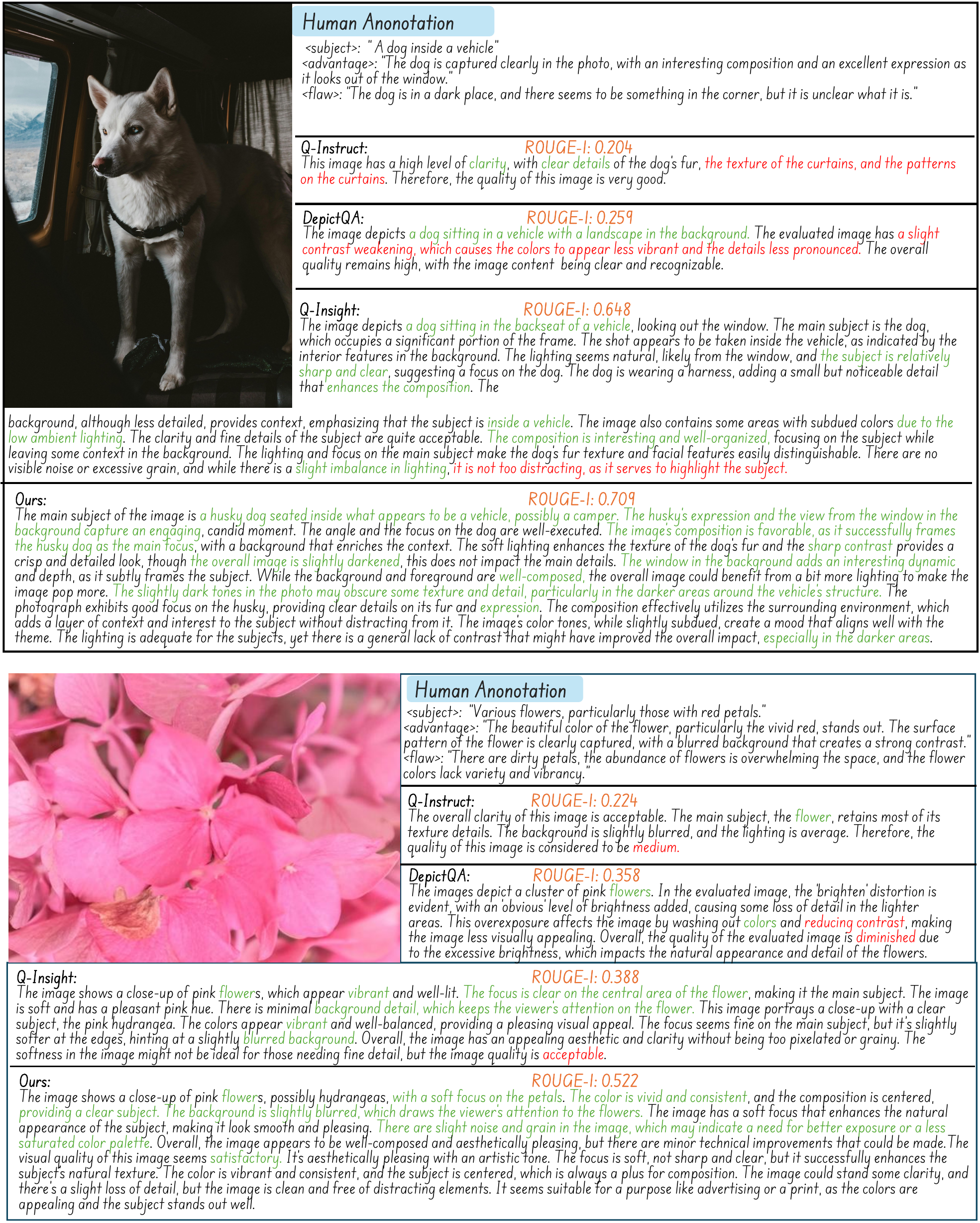}
    \caption{
\textcolor[rgb]{0,0.6,0}{Green text} indicates reasoning parts consistent with human annotations,
while \textcolor[rgb]{0.8,0,0}{red text} highlights inconsistencies. Please zoom in for details. In the upper subfigure, both our model and Q-Insight~\cite{q-insight} detect the presence of dark regions; 
however, Q-Insight~\cite{q-insight} incorrectly concludes that they do not affect image quality, 
whereas our model—consistent with human judgment—correctly identifies them as quality-degrading factors. 
In the both subfigures, our model produces more fine-grained and human-aligned descriptions, 
using terms such as “expression’’ and “petal’’ while accurately identifying quality-reducing factors, 
all of which align with human annotations.
}
    \label{fig:s2}
\vspace{-15pt}
\end{figure*}

\begin{figure*}[!h]
    \centering
    \includegraphics[width=1.0\textwidth]{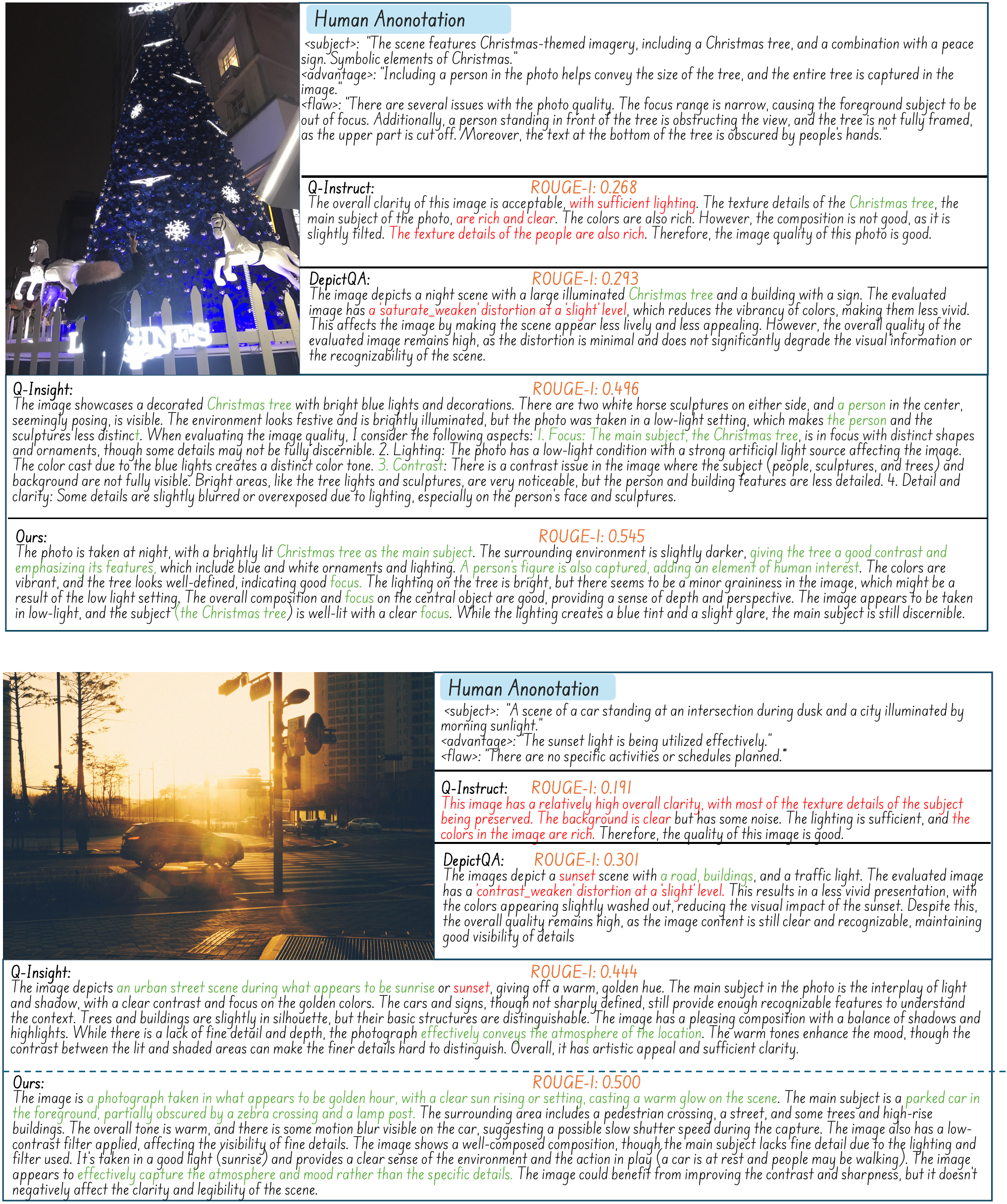}
    \caption{
\textcolor[rgb]{0,0.6,0}{Green text} indicates reasoning parts consistent with human annotations,
while \textcolor[rgb]{0.8,0,0}{red text} highlights inconsistencies. Please zoom in for details. Q-Instruct~\cite{q-instruct} and DepictQA~\cite{depictqa} exhibit strongly template-like outputs, and for Q-Instruct~\cite{q-instruct}, 
even clear perceptual mistakes do not affect the final score—suggesting the absence of genuine reasoning. 
Q-Insight~\cite{q-insight} also displays a procedural, pattern-driven reasoning style that diverges from human judgments, 
as seen in the upper subfigure. 
In contrast, our model not only captures fine-grained perceptual cues but also incorporates 
higher-level conceptual factors such as overall atmosphere. 
These characteristics lead to reasoning and descriptions that more closely align 
with human perception and expression.
}
    \label{fig:s3}
\vspace{-15pt}
\end{figure*}

In Figure.~\ref{fig:s2} and \ref{fig:s3}, we compare the perception-reasoning behaviors of Q-Instruct~\cite{q-instruct}, DepictQA~\cite{depictqa}, Q-Insight~\cite{q-insight}, and our model (detailed) against human annotations. These examples reveal that SFT-based models (Q-Instruct~\cite{q-instruct} and DepictQA~\cite{depictqa}) tend to generate highly template-like responses, and these text contributes little to the interpretability of the quality prediction. Even when their reasoning is incorrect, the final score often remains unchanged, indicating that these models do not actually rely on the reasoning process they present. To address this issue, our training strategy employs an RL–based \textit{reasoning reward} rather than SFT supervision. 

Compared with Q-Insight~\cite{q-insight}, our model exhibits stronger human-aligned behavior at both the perception and reasoning levels.
At the perception level, our model attends to a wider range of quality-relevant factors that humans naturally notice—such as near-field blur and fine-detail loss (e.g., Fig.~\ref{fig:s3}).
At the reasoning level, our model more accurately interprets how these perceptual cues affect overall image quality.
For instance, as shown in Fig.~\ref{fig:s2}, humans and our model recognize that dark regions degrade perceived quality, whereas Q-Insight~\cite{q-insight} incorrectly concludes that they have no influence.
These observations suggest that, under similar RL-based training frameworks, our model achieves a higher degree of perception–reasoning alignment with human judgment.

\end{document}